\documentclass[12pt,a4,notitlepage,english]{llncs}
\usepackage[T1]{fontenc}
\usepackage[utf8]{inputenc}
\usepackage[english]{babel}
\usepackage{alltt}
\usepackage{paralist}
\usepackage{listings}
\usepackage{hyperref}
\usepackage{times}
\usepackage{color}
\usepackage{amssymb}
\usepackage{tikz}

\urldef{\mails}\path|{ramona.enache,inari.listenmaa,prasanth.kolachina}@cse.gu.se| 
\usepackage{gb4e}
\noautomath

\title{Handling non-compositionality in multilingual CNLs}
\author{Ramona Enache \and Inari Listenmaa \and Prasanth Kolachina}

\institute{University of Gothenburg\\
\mails}

\titlerunning{Handling non-compositionality in multilingual CNLs}

\authorrunning{Handling non-compositionality in multilingual CNLs}

\begin{document}
\maketitle


\begin{abstract}

In this paper, we describe methods for handling \textit{multilingual} non-compositional
constructions in the framework of GF. 
We specifically look at methods to detect and extract
non-compositional phrases from parallel texts 
and propose methods to handle such constructions in GF grammars. 
We expect that the methods to handle non-compositional constructions
will enrich CNLs by providing more flexibility in the design of
controlled languages. 
We look at two specific use cases of non-compositional constructions:
a general-purpose method to detect and extract multilingual multiword
expressions and a procedure to identify nominal compounds in German. 
We evaluate our procedure for multiword expressions by performing a 
qualitative analysis of the results.
For the experiments on nominal compounds, we incorporate the detected
compounds in a full SMT pipeline and evaluate the impact of our method in machine translation process. 


\end{abstract}

\section{Introduction}
\label{section:Introduction}

The work describes a series of methods used to enrich multilingual
CNLs written in the grammar formalism GF (Grammatical
Framework)\cite{ranta2011book} with multilingual multiword
expressions (MMWEs). This aims to give a better separation between
compositional and non-compositional constructions in GF applications
and a better understanding on representing MMWEs in GF. We present two
new GF modules: one for constructions in a multilingual setting, and one specifically for German compound nouns.

We are targeting cases where translation equivalents have different
syntactic structure: this covers pairs such as English--French (\emph{apple juice},
\emph{jus de pommes} `juice of apples') and English--Finnish
(\emph{kick the bucket}, \emph{heittää henkensä} `throw one's
life'). Only the latter pair contains monolingually non-compositional
structure, i.e. having an interpretation that cannot be inferred from
the components, but we consider both of them as MMWEs, due to the non-compositionality of translation.

We propose a solution to this, that relies on prior analysis of the
domain, since GF applications are normally developed starting from
positive examples covering the domain \cite{BestPractice}. We start
from a parallel corpus describing the scope of the grammar and
identify MMWEs in order to add them to the grammar as special
constructions. 

A special case of MMWEs, which we treat separately is that of nominal
compounds in German. The need for a multilingual lexicon of such
compounds and their translations originated from the use of GF in
machine translation \cite{EAMT}, \cite{D5.3}. This use case is of
particular interest, since it is easier to evaluate---both in terms
of precision and recall of the method, and in terms of impact in the
machine translation process.

This paper is structured as follows:
Section \ref{section:RelatedWork} describes the background and related
work;
Section \ref{section:Implementation} describes the implementation of
the general MWE detection and compound detection methods; 
Section \ref{section:Evaluation} describes a preliminary evaluation, 
and finally 
Section \ref{section:FutureWork} describes future work.

\section{Background and Related Work}
\label{section:RelatedWork}
\subsection{Grammatical Framework}

GF (Grammatical Framework) is a grammar formalism particularly fit for
multilingual natural language applications. In the recent years, it
has been used extensively for developing (multilingual) CNLs, such as  
the in-house implementation of Attempto Controlled English \cite{ranta2009cnl}, and
domain-specific applications for mathematical exercises \cite{saludes2011mathematics}, \cite{saludes2012sage}, \cite{saludes2012multilingualMathAssistant}, 
speech-based user interfaces \cite{kaljurand2012cnl1}, tourist
phrases \cite{ranta2010cnlphrasebook}, business models
\cite{beinformed} and cultural heritage artifacts \cite{CH2}, \cite{CH3}.

Applications written in GF are represented by their {\it abstract
  syntax}, which models the semantics of the domain in a
language-independent fashion, and a number of {\it concrete syntaxes},
mapping the semantics to a number of target languages, most commonly
natural languages. 

The difficulty when dealing with compositional and non-compositional
constructs in GF arises, in fact, from the multilingual character of
the applications. It is of particular difficulty to design the
abstract syntax in a way that accommodates all the concrete syntaxes,
without the need for further change. As a potential solution to this,
there has been work done on deriving the abstract syntax from an
existing ontology \cite{angelov-enache-2010} or FrameNet
\cite{Normunds}, \cite{Normunds2014LREC}. However, such resources are not always available.

\subsection{Multiword expressions}

There is a significant body of research on MWEs, ranging from
classification \cite{baldwin-handbook10}, linguistic analysis
\cite{PainInNeck} to methods to detect MWEs (for both monolingual
\cite{MWEDetect1}, \cite{MWEDetect2} and
multilingual settings \cite{MWEDetect3}, \cite{RandomMWE1}, \cite{RandomMWE2}) and
evaluation measures for these methods \cite{MWEEval1}.

Following the MWE taxonomy from \cite{baldwin-handbook10} into fixed,
semi-fixed and syntactically flexible expressions, we note that
applying the same scale to MMWEs, it is the semi-fixed and
syntactically flexible constructions that are most effectively
representable in GF. The reason is that GF allows for generalisations
in terms of arguments (for relational MWEs, such as transitive verb
phrases), declension forms and topicalisation in the sentence.

\section{Methods for MMWE Extraction}
\label{section:Implementation}

\subsection{General MMWE candidate extraction}


The algorithm for general MMWE extraction parses a pair of sentences
$(X, Y)$ with a wide-coverage GF grammar, often resulting in multiple
parse trees for each sentence. Then it compares all pairs of trees \\
\{$(x, y) \; | \; x \in parse(X), y \in parse(Y)$\}, and if no identical trees are
found, the phrases are candidates for containing BMWEs. 

Part of the test material was not parsed by the regular GF grammar. To
add robustness, we used a new chunking
grammar\footnote{https://github.com/GrammaticalFramework/GF/blob/master/lib/src/experimental/Chunk.gf}
for the language pair English--Swedish. French and
German didn't have the chunking grammar implemented, so for pairs
including them, we used robust parsing in GF
\cite{angelov2011mechanics}, \cite{angelovljunglof2014}. 
With the chunking grammar, the trees kept their local structure better,
whereas the robust parser resulted in flatter
structure, making the distance to any well-formed tree high.
Thus these sentences were always reported as BMWE candidates.
For our small test set, this wasn't a problem, but for future work, a
fallback for partial trees should be considered, e.g. one that
translates the sentences both ways and calculates the word error rate.




We used material from two sources. First, we took 246 sentences from
the Wikitravel phrase collection\footnote{http://wikitravel.org/en/List\_of\_phrasebooks} in English, German, French and Swedish.
The material consists of sentences such as asking for direction or
expressing needs, in various language pairs of which other is English. 
For another type of text, we took the 61--sentence short story ``Where is James?'', 
from the website UniLang\footnote{http://www.unilang.org/}, 
 which contains free material for language learning. 
 In total our test set was 307 sentences, functioning mostly as a proof of concept.





After running the experiments, we found various MMWE candidates in all
language pairs.
We added relevant new findings to the GF multilingual dictionary, some
replacing the old translations, some as new lexical items.
However, the majority of the candidates were predicates that span over a larger
structure, and couldn't be covered just by lexicon---instead, we added
them to a new module, called \emph{Construction}
(see Figure~\ref{fig:Constructions}).

The module is, in the spirit of construction grammar, between syntax
and lexicon. Instead of applying to categories in general, most of the
functions in the module are about particular predicates which are
found to work differently in different languages. The purpose of the
module is hence not so much to widen the scope of string recognition,
but to provide trees that are abstract enough to yield correct
translations. It is being developed incrementally, but we envision
being able to develop the module in a more systematic manner by
employing data-driven methods, such as extracting constructions from a
treebank.

\begin{figure}[t]
\begin{quote}
\scriptsize
\begin{verbatim}
weather_adjCl : AP -> Cl ;        -- it is warm / il fait chaud (Fre)
n_units_AP    : Card -> CN -> A -> AP ;  -- x inches long
glass_of_CN   : NP -> CN ;        -- glass of water / lasillinen vettä (Fin)
where_go_QCl  : NP -> QCl ;       -- where did X go / vart gick X (Swe)
\end{verbatim}
\normalsize
\end{quote}
\caption{Example of constructions}
\label{fig:Constructions}
\end{figure}




\subsection{GF lexicon of compound words}
\label{section:GFcompounds}

A substantial part of the work on MWEs involved the detection and
representation of compound words in GF. The motivation for this
lies in the need to improve GF-driven machine translation from English
into German, especially in the bio-medical domain \cite{ramona2013thesis}.

The goal is to extract pairs consisting of German compound words and
their English translations from parallel corpus, to syntactically analyse
the compound and to build a GF representation of the pair, which will be
added to a compound lexicon. 
Because the most frequent such compound words are
nominals \cite{baldwin-handbook10}, we consider them as the use case of our method.

The method relies on a GF resource describing rules for nominal
compounding. 
The following rules describe three types of compounding: first one
with the modifier in nominative, second one with the morpheme `s' in
the end (\emph{Lebensmittel} `life-s-means') and third one
with the ending `en' (\emph{Krankenwagen} `sick-en-vehicle').

\begin{figure}[t]
\begin{quote}
\scriptsize
\begin{verbatim}
fun ConsNomCN : N -> CN -> CN ; 
fun Cons_sCN : N -> CN -> CN ;
fun Cons_enCN : N -> CN -> CN ;
\end{verbatim}
\normalsize
\end{quote}
\caption{Example of compounds}
\label{fig:GermanCompounds}
\end{figure}

The basic procedure is the following: 
\begin{itemize}
\item we extract candidate pairs, which fulfil the following criteria:
   \begin{itemize}
      \item their probability is above a confidence threshold
      \item the English part parses as an NP in GF
      \item the German part is composed of one word
   \end{itemize}
\item we employ a greedy algorithm to split the German word into a 
  number of lexical items from the German monolingual dictionary from
  GF (based on Wiktionary), based on the German compound grammar
  described above; 
  we select the split which employs the
  least number of tokens
\item we add the pair of GF trees to a lexical resource for compounds
\end{itemize}

In our experiments, phrase translations extracted from a English-German parallel corpus~\cite{koehn:2005europarl}
are used to detect possible nominal compounds in German. 
For practical reasons, we restrict the set of possible phrase translations to phrases determined to be
\textit{constituents} in the parse tree for the English sentence by a constituency parser~\cite{klein:2003stanford}.
This restricts the amount of noise in the translation memories, where \textit{noise} is defined as a
pair of random sequence of words in English and German that are seen together in the translations. Furthermore, 
we restrict our interest to entries that are labelled as noun phrases
by the parser.

%




\section{Evaluation}
\label{section:Evaluation}

\subsection{Evaluation of general MMWE extraction}

As a tentative evaluation for the general MMWE extraction method, we
used the results of the language pair English--Swedish and did qualitative analysis of the findings.
We chose Swedish, because it had the best grammar coverage out of the
languages we tested; the results for French and German were poorer,
due to the flat structure of trees from robust parsing. 
The chunking grammar made it possible to compare trees even when one has a
complete parse and other not, since the well-formed sentence can also be
expressed as chunks.

\begin{table}[h]
\begin{center}
\begin{tabular}{|r|r|c|}
\hline
\multicolumn{2}{ |l| }{Not MWE candidates} & 92 \\ \hline
\multicolumn{2}{ |l| }{MWE candidates}  & 215  \\ \hline
\multicolumn{2}{ |r| }{False positives}  & 44  \\ 
\multicolumn{2}{ |r| }{Lexical MWEs}     & 29  \\ 
\multicolumn{2}{ |r| }{Predicates}       & 142 \\ \hline
\multicolumn{2}{ |l| }{\textbf{All sentences}} & \textbf{307} \\ \hline

 \end{tabular}
\end{center}
\caption{General MMWE extraction}
\label{table:GeneralMMWE}
\end{table}

Table~\ref{table:GeneralMMWE} shows the results of the analysis.
Of the 307 sentences in English and Swedish, we found 215 candidates,
of which 44 were considered false positives, due to parsing problems.
For the algorithm to recognise two sentences as identical, it needs to
have parsed them properly, so we did not get false negatives.

Out of the remaining 171 candidates, we classified 29 to
be lexical MWEs, such as English \emph{locker} vs. Swedish
\emph{låsbart skåp}
`lockable closet', or \emph{hide from} vs. \emph{gömma
  sig för} `hide \textsc{refl} for'. Not all of them were one-to-many; in
11 cases it was just a question of similar words, such as
\emph{little} and \emph{small} used in the parallel sentences. 

142 candidates were predicates that span over a larger structure. 
The expressions could be classified to the following subcategories:
\begin{inparaenum}[a\upshape)]
\item greetings;
\item weather expressions;
\item time expressions;
\item money;
\item units of measurement, containers;
\item spatial deixis.
\end{inparaenum}


These expressions are non-compositional due to different factors:
e.g. greetings and weather expressions are highly idiomatic, fixed
phrases. Other cases, such as units, are less rigid: a certain
semantic class of words appears in structures like \emph{glass of NP},
which work differently in different languages. For example, Swedish uses no
preposition,  Finnish uses a special form \emph{glassful}. 
Since adding a general rule for \emph{NP of NP} would be overgenerating, we
added these constructions separately for each container word
(e.g. \emph{glass, bottle, cup, bucket}). 

An example of spatial deixis is the correspondence of direction adverbs
between languages: e.g. the word \emph{where} in the sentence
\emph{where did X go}  should be translated in German to \emph{wohin}
`where to' instead of \emph{wo} `where in'; same with \emph{here} and
\emph{there}. We added these constructions as combinations of a motion verb and a direction adverb. 

Finally, a number of the 142 phrases were correctly
recognised as containing a differing subtree, but we judged the
difference not to be general enough to be added as a construction.
For example, sentence (1) from the short story has the
auxiliary verb \emph{can} in the English version and not in the Swedish,
and the adverb \emph{tydligt} means `clearly, distinctly'. 
While not general enough for the construction module, results like this
could still be useful for some kind of application grammar; the method
correctly recognises them, as long as the sentences are fully parsed.

\begin{exe}
\ex
\gll Hon hör det tydligt nu (Swe) \\
`she hears it clearly now' \\
\glt She can hear it well now (Eng)
\end{exe}

\subsection{Evaluation of German nominal compounds}

We evaluated the German nominal compounds detected by our algorithm
based on their utility in the task of machine translation. 
In this experiment, we provided the detected nominals as possible
dictionary items to an SMT pipeline and extracted a translation memory
from a news domain corpora augmented with the nominal compounds. We evaluated the improvements in translation quality after augmenting the translation memories
with these nominal compounds. Translation quality is evaluated in
terms of BLEU score, a standard metric used in evaluating performance
of MT systems. Table~\ref{tbl:mosesscores} shows the BLEU scores
obtained from two different SMT systems, a baseline system and the
same system using the translation memory augmented with nominal
compounds. The BLEU scores are reported on standard test datasets used
in the evaluation of SMT systems.\footnote{The datasets can be found at \url{http://www.statmt.org/wmt14/translation-task.html}. 
We use the newstest2011 and newstest2012 datasets in our experiments.}

The improvement gained by using this simple method suggests that a
proper handling of MWEs could improve the BLEU scores in an even more
significant manner, by taking advantage of the full power of the GF
representations, mainly by aligning all declension forms of MWEs and
adding them to the translation memories.

\begin{table}
\begin{center}
\begin{tabular}{|l|c|c|}
\hline
SMT system & newstest2011 & newstest2012 \\ \hline
Baseline & 11.71 & 11.64  \\
+Compounds & 11.83 & 11.96 \\
\hline
\end{tabular}
\end{center}
\caption{BLEU scores obtained from the SMT systems}
\label{tbl:mosesscores}
\end{table}


%

\section{Future work}
\label{section:FutureWork}




As GF has proven to be a reliable environment for writing multilingual
CNLs and compositionality is a known problem of such applications, 
our method to isolate non-compositional constructions would be a great
aid for the development of GF grammars, if it were applied on more
domains and language sets. In this manner, one could also asses the
generality of the method, both in terms of languages and
types of constructions, more clearly. 

For the purpose of aiding the development of GF domain grammars, we
are also considering a combination between our method and the related efforts of constructing
multilingual FrameNet-based grammars \cite{Normunds}, \cite{Normunds2014LREC}.

Regarding the use of MWE in machine translation, one can consider
integrating the GF resources in a more meaningful manner, by not just
aligning the basic forms, but also the declension forms. The MWE
resources could also be helped to improve the existing GF-driven hybrid
translation systems \cite{ramona2013thesis}.

Last, but not least, as our initial experiments have shown a rather large
number of false positives, we aim to develop specific pre-processing
methods to address this issue. A boost in accuracy would lead to a
decrease in the size of the initial resources that are automatically
created and reduce the effort for evaluation. A possible solution
would be comparing the shape of the parse trees, in order to asses
differences in the constructions. 

In conclusion, our work represents the first step in handling
non-compositional constructions in multilingual GF
applications. The methods are still under development, but they still
highlight the significant advantages that the feature brings, both
to general CNLs written in GF and to large translation systems.   
 









\section*{Acknowledgements}

The authors would like to thank Koen Claessen and Aarne Ranta for
their input on both the methods developed and this paper. Moreover, we
would like to thank Víctor Sánchez-Cartagena for the fruitful
discussion on a previous version of the MWE detection algorithm and
the ideas on how to implement it for parallel free text.

We also want to thank the Swedish Research Council for financial 
support under grant nr. 2012-5746 (Reliable Multilingual Digital 
Communication: Methods and Applications).


\bibliography{compounds,smt}
\bibliographystyle{plain}
\end{document}